\documentclass{l4dc2026}

\title[GCImOpt: Learning efficient goal-conditioned policies by imitating optimal trajectories]{\textsc{GCImOpt}: Learning efficient goal-conditioned policies by imitating optimal trajectories}
\usepackage{times}

\usepackage{siunitx}
\usepackage{bm}
\usepackage{physics}
\usepackage[thinc]{esdiff}
\usepackage{microtype}
\usepackage{inconsolata}
\usepackage{booktabs}

\SetKwComment{Comment}{/* }{ */}

\author{%
 \Name{Jon Goikoetxea} \Email{Jon.Goikoetxea-Macua@etu.univ-grenoble-alpes.fr}\\
 \addr Department of Statistics, Mathematics and Computer Science, Public University of Navarre\\
 \addr Ensimag, Institut National Polytechnique de Grenoble, Université Grenoble Alpes
 \AND
 \Name{Jesús Palacián} \Email{palacian@unavarra.es}\\
 \addr Department of Statistics, Mathematics and Computer Science, Public University of Navarre\\
 \addr Institute for Advanced Materials and Mathematics (InaMat$^2$)
}

\begin{document}

\maketitle

\begin{abstract}%
Imitation learning is a well-established approach for machine-learning-based control.
However, its applicability depends on having access to demonstrations, which are often expensive to collect and/or suboptimal for solving the task.
In this work, we present \textsc{GCImOpt}, an approach to learn efficient goal-conditioned policies by training on datasets generated by trajectory optimization.
Our approach for dataset generation is computationally efficient, can generate thousands of optimal trajectories in minutes on a laptop computer, and produces high-quality demonstrations.
Further, by means of a data augmentation scheme that treats intermediate states as goals, we are able to increase the training dataset size by an order of magnitude.
Using our generated datasets, we train goal-conditioned neural network policies that can control the system towards arbitrary goals.
To demonstrate the generality of our approach, we generate datasets and then train policies for various control tasks, namely cart-pole stabilization, planar and three-dimensional quadcopter stabilization, and point reaching using a 6-DoF robot arm.
We show that our trained policies can achieve high success rates and near-optimal control profiles, all while being small (less than \num{80000} neural network parameters) and fast enough (up to more than $6000\times$ faster than a trajectory optimization solver) that they could be deployed onboard resource-constrained controllers.
We provide videos, code, datasets and pre-trained policies under a free software license;
see our project website \url{https://jongoiko.github.io/gcimopt/}.

\end{abstract}


\begin{figure}[h!]
    \centering
    \includegraphics[width=0.8\textwidth]{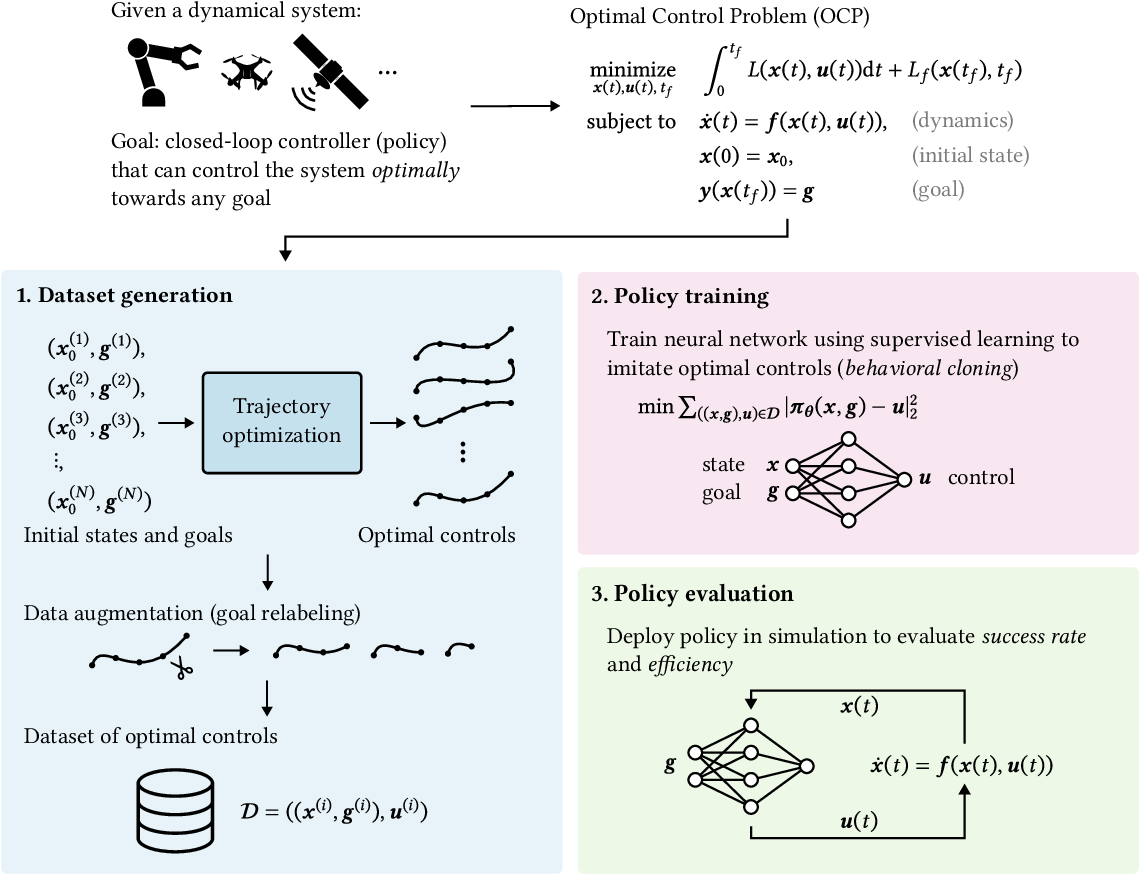}
    \caption{Schematic summary of \textsc{GCImOpt}, a simple method to train and evaluate near-optimal goal-conditioned policies.
    A policy $\bm{\pi}_\theta$, parameterized by a neural network, is trained by supervised learning on a large dataset of trajectories generated by trajectory optimization.
    The policy is then evaluated in simulation to measure its success rate and efficiency.}
    \label{fig:method}
\end{figure}

\section{Introduction}

Optimally controlling a dynamical system enables the accomplishment of a task while minimizing a given cost measure.
However, for many dynamical systems, finding an optimal closed-loop controller or policy is very difficult or impossible.
For this reason, trajectory optimization \citep{diehl_numerical_2011,betts_practical_2010} is often used in practice, for example as part of model predictive control (MPC) \citep{rawlings_model_2017}:
the controller computes open-loop optimal controls in a loop at high frequency, executing at each step the first segment of the optimal control.
Even though MPC allows the design of near-optimal closed-loop controllers, solving optimization problems at high frequency makes it computationally expensive.

In addition to obtaining controllers that are efficient and have low computational cost, we are interested in having them be \emph{general}:
a general policy would be able to accomplish multiple control tasks on the system without having to modify the policy itself.
This can be obtained by passing a representation of a \emph{goal} to the controller so that its output is conditioned both on the current system state and the desired goal.
Generalization is a central problem in machine learning (ML), since we aim to train models that perform well on the entire data distribution, even when faced with previously unseen data.
Thus, a data-driven method can be effective to obtain near-optimal goal-conditioned policies that have low computational cost at inference:
using trajectory optimization to generate a large dataset of optimal controls, we can train a small neural network to approximate the optimal controller and generalize to arbitrary goals.

\subsection{Related work}

\paragraph{Goal conditioning} Goal-conditioned policies have been extensively studied in the context of reinforcement learning \citep{kaelbling_learning_1993,schaul_universal_2015,liu_goal-conditioned_2022}.
A fundamental advantage of goal conditioning is that trajectories can be \emph{relabeled} to have their goals be different to the originally intended goals \citep{andrychowicz_hindsight_2017}, achieving greater sample efficiency than direct RL on goal-conditioned tasks.
In spite of this, GCRL approaches often require careful reward shaping and online environment interaction, introducing the challenge of exploration;
these difficulties can be avoided by resorting to imitation learning.
In goal-conditioned imitation learning (GCIL), we require datasets where the expert solves multiple tasks, which may then be used to speed up training RL agents \citep{ding_goal-conditioned_2019}, use the expert's intentions as goals \citep{codevilla_end--end_2018} or imitate multimodal experts with expressive policy classes \citep{reuss_goal-conditioned_2023}.
Additionally, GCIL may be scaled to use massive and heterogeneous datasets, enabling the training of general models that work on various robot morphologies \citep{shah_gnm_2023}.
GCIL also relaxes the need for expert behavior in the dataset, meaning that suboptimal, uncurated data can still be leveraged for policy learning \citep{lynch_learning_2020,ghosh_learning_2019}.

\paragraph{Imitation learning on optimal trajectories} Trajectories obtained by trajectory optimization can be used as expert data for imitation learning.
Earlier work makes a careful consideration of the interplay between policy learning and trajectory optimization, e.g.\ penalizing the optimized trajectories' deviation from the policy's outputs \citep{mordatch_combining_2014,levine_guided_2013,zhang_learning_2016,kahn_plato_2017} or leveraging the structure of system dynamics to ensure the policy's stability \citep{da_combining_2019}.
In recent years, there has been a renewed interest in learning from optimal trajectories computed offline, with special emphasis on aerospace applications \citep{sanchez-sanchez_optimal_2016,sanchez-sanchez_real-time_2018,rubinsztejn_neural_2020,sprague_learning_2020}.
A notable characteristic of these works is the simplicity of the training setup, since policies are directly trained using behavioral cloning rather than using a more complex algorithm such as \textsc{DAgger} \citep{ross_reduction_2011}.
\citet{tailor_learning_2019} attribute this advantage to low input dimensionality and large dataset sizes, which minimize the impact of compounding errors in the policies' predictions.
Policies trained using this method have been deployed in real quadcopters \citep{ferede_end--end_2024}, where they have been shown to perform comparably to differential flatness-based controllers.
Although behavior cloning of policies on optimal trajectories has been extensively studied, little work has been done on extending the generality of said policies by means of goal conditioning.

\subsection{Contributions}

The contributions of this work are the following:
\begin{itemize}
  \item We present \textsc{GCImOpt}, a simple method to train and evaluate near-optimal goal-conditioned policies on datasets generated by trajectory optimization (Fig.\ \ref{fig:method}).
  Unlike GCRL methods, our method does not require reward shaping or online environment interaction, since training is done on optimal trajectories computed offline.
  We expand on previous work by considering the effects of goal conditioning, thus training policies that can generalize to arbitrary goals while remaining near-optimal.
  We use a direct method for trajectory optimization, thus avoiding the need to analytically construct optimality conditions for each considered control problem;
  in addition, we make dataset generation fast by using the \textsc{Fatrop} solver \citep{vanroye_fatrop_2023}, which is purpose-built for optimal control applications.
  \item To demonstrate the generality of \textsc{GCImOpt}, we evaluate it empirically on four different simulated dynamical systems.
  We thus also evaluate the impact of the system's dimensionality and nonlinearity on the trained policies' performance.
  We show that \textsc{GCImOpt} policies can achieve high success rates and near-optimal control profiles.
  Moreover, we show that \textsc{GCImOpt} policies can achieve speedups of $97\times$ to $6278\times$ versus the fast \textsc{Fatrop} solver, suggesting that they could surpass the control frequency of MPC controllers.
  \item We release an implementation of the method, the code of the experiments, the generated datasets and the trained policies under a free software license.
  See our project website \url{https://jongoiko.github.io/gcimopt/} for access to these resources.
\end{itemize}
In the next section, we formalize the problem of finding optimal goal-conditioned policies and recall the basic notions of numerical trajectory optimization.
We detail the steps of our method in Sec.\ \ref{sec:gcimopt}, and present the experimental evaluation and its results in Sec.\ \ref{sec:exp}.
Finally, we conclude by discussing the method's advantages and limitations and describing ideas for future work.
Additional experiment details and tables are provided in Appendix~\ref{sec:hyperparams}.

\section{Preliminaries: Closed-loop optimal control with goals}

In order to generate high-quality training trajectories that leverage the system's dynamics, we use numerical trajectory optimization.
This is a class of methods for solving optimal control problems (OCPs) specifying a control task to be solved while minimizing some measure of cost.
We consider continuous-time dynamical systems which can be described by ODEs of the form $\bm{\dot x}(t) = \bm{f}(\bm{x}(t), \bm{u}(t))$, where $\bm{x}(t) \in \mathbb{R}^{n_x}$ and $\bm{u}(t) \in \mathbb{R}^{n_u}$ denote the system's state and the control at time $t$ respectively.

In this work, we solve OCPs that are constrained to a \emph{goal} being reached at the end of the trajectory, which may be a system state or some other representation derived from a state.
Thus, we assume that there is a \emph{state-to-goal mapping} $\bm{y}: \mathbb{R}^{n_x} \to \mathbb{R}^{n_g}$ that, given a state, returns the goal reached by the system being in that state.
We aim to approximate the optimal goal-conditioned closed-loop controller (or \emph{policy}) for the given OCP, which we define as
\begin{subequations}\label{eq:optimal-policy}
\begin{align}
    \bm{\pi}^*(\bm{x}_0, \bm{g}) = (\arg & \min_{\bm{u}(t), t_f} \int_{0}^{t_f} L(\bm{x}(t), \bm{u}(t)) \dd t + L_f(\bm{x}(t_f), t_f))(0)\label{eq:cost-functional} \\
    \text{subject to} \quad & \bm{\dot x}(t) = \bm{f}(\bm{x}(t), \bm{u}(t)), \quad t \in [0, t_f],\\
    & \bm{x}(0) = \bm{x}_0,\label{eq:initial-state-constraint} \\
    & \bm{h}(\bm{x}(t), \bm{u}(t), t) \le \bm{0}, \quad t \in [0, t_f],\label{eq:path-constraint}\\
    & \bm{y}(\bm{x}(t_f)) = \bm{g}\label{eq:goal},\\
    & t_{\min} \le t_f \le t_{\max}.
\end{align}
\end{subequations}
Thus, the optimal policy's output is the solution to a constrained OCP:
the control minimizes the OCP's cost functional (Eq.\ \ref{eq:cost-functional}) with the system starting at state $\bm{x}_0$ (Eq.\ \ref{eq:initial-state-constraint}) and demanding the goal $\bm{g}$ to have been reached after a time $t_f$ (Eq.\ \ref{eq:goal}).
The cost functional is composed of a \emph{Lagrange term} where a function $L$ is integrated over the trajectory duration, and a \emph{Mayer term} which may incur additional cost at the end of the trajectory, as defined by the function $L_f$.
In Eq.\ \ref{eq:cost-functional}, we use the notation $(\arg \min_{\bm{u}(t), t_f} \cdots) (0)$ to denote that, given the optimal open-loop control function $\bm{u}(t)$ defined on $t \in [0, t_f]$, the closed-loop policy outputs the control at the current time $t=0$, that is $\bm{u}(0)$.
We use the path constraint (Eq.\ \ref{eq:path-constraint}) to incorporate state and actuator limits, and we treat the trajectory duration $t_f$ as a decision variable constrained to a range $[t_{\min}, t_{\max}]$.

We solve OCPs of the given form numerically using a \emph{direct} method.
This is in contrast to \emph{indirect} methods such as those based on Pontryagin's Minimum Principle \citep{pontryagin_mathematical_1962}, which require analytically constructing the OCP's necessary and sufficient conditions for optimality.
In particular, we use direct multiple shooting \citep{bock_multiple_1984} to \emph{transcribe} the OCP to a non-linear program (NLP) in a finite number of decision variables, which is then solved numerically using an NLP solver.
For a detailed discussion of numerical trajectory optimization, we refer to \citet{diehl_numerical_2011,betts_practical_2010,betts_survey_1998}.

\section{Learning efficient goal-conditioned policies by imitating optimal trajectories}\label{sec:gcimopt}

We aim to train a goal-conditioned policy $\bm{\pi}_\theta$ to generalize to arbitrary system states and goals, approximate the optimal policy for the given OCP (defined in Eq.\ \ref{eq:optimal-policy}) and be computationally efficient.
To that end, we generate optimal input-output pairs by trajectory optimization (where the inputs are state-goal pairs and the outputs are the corresponding optimal controls).
This yields a large dataset $\mathcal D = \{(\bm{x}^{(i)}, \bm{g}^{(i)}), \bm{u}^{(i)}\}_{i=1}^{N_{\mathcal D}}$ of training samples that we can use to train $\bm{\pi}_\theta$ by supervised learning.
For simplicity, in this work we assume deterministic environment dynamics and thus train deterministic policies.

\subsection{Dataset generation}

We generate a large dataset of optimal trajectories, each of which has an imposed initial state and goal.
For each trajectory, we sample a pair of states from a distribution $\mathcal T$ over $\mathbb{R}^{n_x} \times \mathbb{R}^{n_x}$, which we call the \emph{task distribution}:
from a pair of states $(\bm{x}_0, \bm{x}_f) \sim \mathcal T$, the first state $x_0$ is the initial state in the trajectory, while the second $\bm{x}_f$ is used to generate the goal $\bm{g} = \bm{y}(\bm{x}_f)$.
We then solve the corresponding trajectory optimization problem using direct multiple shooting (DMS) \citep{bock_multiple_1984}.
In DMS, the control and state trajectory over time is discretized into a finite number of decision variables $\bm{x}_i, i = 0, 1, \cdots, N$ and $\bm{u}_i, i = 0, 1, \cdots, N-1$.
Each discretized time instant $i$ corresponds to a multiple shooting \emph{grid point}.
The dynamic feasibility of the finite-dimensional approximation is ensured by introducing \emph{gap-closing} constraints of the \emph{explicit} form $\bm{x}_{i+1} = F(\bm{x}_i, \bm{u}_i)\; \forall i \in \{0, 1, \cdots, N-1\}$, where $F$ numerically integrates system dynamics over time\footnote{Since DMS requires system dynamics to define a well-posed initial value problem (IVP) that can be solved numerically, the dynamics function $\bm{f}$ is assumed to be locally Lipschitz in $\bm{x}(t)$ and continuously differentiable in $\bm{x}(t)$ and $\bm{u}(t)$.}.
In this work, we use a 4th order Runge-Kutta (RK4) for the numerical integration of the system dynamics and cost functional.
We choose multiple shooting as the transcription method (as opposed to e.g.\ collocation) because the \textsc{Fatrop} optimizer expects the discretized dynamics to be written in the explicit form.
From a successfully optimized trajectory, we recover the states and controls $(\bm{x}_i, \bm{u}_i)$ directly from the multiple shooting grid points, for $0 \le i < N$.
As for the initial guess provided to the NLP solver, we initialize all control variables $\bm{u}_i$ to 0, linearly interpolate the states $\bm x_i$ between $\bm{x}_0$ and $\bm{x}_f$, and set $t_f = (t_{\min} + t_{\max}) / 2$.

\paragraph{Dataset augmentation by goal relabeling}
To multiply the number of training samples obtained from the dataset generation process, we use a \emph{goal relabeling} or \emph{hindsight relabeling} technique \citep{andrychowicz_hindsight_2017}.
The core idea is that, if $\bm{x}_i$ is an intermediate state in a trajectory, then all previous state-control pairs $(\bm{x}_j, \bm{u}_j), j < i$ successfully control the system towards $\bm{x}_i$;
thus, this \emph{sub-trajectory} may serve as a demonstration for reaching $\bm{g}_i = \bm{y}(\bm{x}_i)$ as a goal.
Hence, we augment the training dataset by taking each trajectory, and for each of its intermediate states $\bm{x}_i, 1 \le i < N$ generating $i$ new input-output pairs $((\bm{x}_j, \bm{y}(\bm{x}_i)), \bm{u}_j)$ for $0 \le j < i$.
This augmentation is not applied to the validation set.

\begin{sloppypar}
\paragraph{Fast and parallel dataset generation}
Since each generated trajectory is independent of the rest, the dataset generation procedure can be easily parallelized.
In addition, instead of using a general-purpose NLP solver such as \textsc{IPOPT} \citep{wachter_implementation_2006}, we use the \textsc{Fatrop} solver \citep{vanroye_fatrop_2023}, which leverages the structure of NLPs resulting from the direct transcription of OCPs;
this enables generating many more optimal trajectories in less time.
We use \textsc{Fatrop} through its CasADi interface \citep{andersson_casadi_2019}.
\end{sloppypar}

\subsection{Policy training}

After generating a large dataset of input-output pairs, we can proceed to train the policy $\bm{\pi}_\theta$ using supervised learning, thus doing behavior cloning (BC) on the optimal demonstrations\footnote{Any other imitation learning algorithm may be used, such as \textsc{DAgger}.
Also, the policy could be parameterized by a more expressive model such as a diffusion model \citep{chi_diffusion_2023,reuss_goal-conditioned_2023}.}.
Given that the inputs to the policies are low-dimensional state and goal representations, we define the policies as simple multi-layer perceptrons (MLPs).
We favor MLP architectures that are smaller and thus faster for training and inference.

Having $\bm{\pi}_\theta(\bm{x}, \bm{g}) = \text{MLP}_\theta(\bm{w})$, where $\bm{w}$ is a vector representing the state-goal pair $(\bm{x}, \bm{g})$, the policy's output should be an $n_u$-dimensional vector (i.e.\ of the dimension of the controls).
We thus simply train $\bm{\pi}_\theta$ to minimize the mean squared error $\text{MSE}(\bm{\pi_\theta}, \mathcal D_\text{train}) = \mathbb E_{((\bm{x}, \bm{g}), \bm{u}) \sim \mathcal D_\text{train}} \big[ |\bm{\pi_\theta}(\bm{x}, \bm{g}) - \bm{u}|_2^2 \big]$, that is, to minimize the deviation between the policy's outputs and the optimal controls obtained by solving the OCP~\ref{eq:optimal-policy}.
Our implementation of the policies and their training is based on the Equinox library \citep{kidger_equinox_2021} which is based on JAX \citep{bradbury_jax_2018}.

\subsection{Policy evaluation}\label{sec:eval}

Once a policy $\bm{\pi}_\theta$ has been trained by imitation learning on $\mathcal D_\text{train}$, we evaluate its success rate and efficiency when controlling the dynamical system.
\citet{tailor_learning_2019} found that a lower mean absolute error (MAE) in imitating the expert controller leads to more performant policies according to the OCP's cost functional;
however, there comes a point where small gains in validation error no longer improve the efficiency of the policy.
Thus, in this section we present our method to empirically evaluate a trained policy, which is heavily inspired by that of \citet{sanchez-sanchez_real-time_2018}.

\paragraph{Checking for successful goal reaching}
For a given goal $\bm{g}$, we simulate the dynamical system where the policy acts as a controller, i.e. $\bm{\dot x}(t) = \bm{f}(\bm{x}(t), \bm{\pi}_{\bm{\theta}}(\bm{x}(t), \bm{g}))$.
In all experiments, we simulate the environment dynamics with $\bm{\pi}_\theta$ as a controller using Dormand-Prince's 7/8 method for numerical integration \citep{dormand_family_1980}, clipping the policy's output to enforce actuator limits.
We define a function $\textsc{GoalReached}: \mathbb R^{n_x} \times \mathbb R^{n_g} \to \{ 0, 1 \}$, which indicates whether the input state reaches the input goal up to a sufficiently small tolerance.
We can then use this function to decide whether a rollout was successful or not in reaching $\bm{g}$:
given the state trajectory $\bm{x}(t)$ obtained in simulation, we take the \emph{completion time} $t^*$ as the earliest time in $[t_{\min}, t_{\max}]$ such that $\textsc{GoalReached}(\bm{x}(t^*), \bm{g}) = 1$;
if no such time exists, then we take the trajectory as unsuccessful, that is, the policy was not able to reach the commanded goal.

\paragraph{Evaluating efficiency}
Assume that the policy successfully reached its goal at completion time $t^*$.
To assess how efficiently it reached the goal with respect to the OCP's cost functional, we compare its achieved cost $J$ (as defined by the cost functional of Eq.\ \ref{eq:cost-functional}) to the optimal cost $J^*$ obtained through trajectory optimization for the same initial state-goal pair.
Since the learned policy is generally suboptimal ($J^* < J$), we quantify its performance using the relative error $\delta_r = 100 \times \frac{J - J^*}{J^*}$, which expresses (as a percentage) how much higher the policy's cost is than the optimal one.
A lower $\delta_r$ indicates a more near-optimal policy.

The evaluation procedure estimates the policy's success rate and efficiency over the full task distribution $\mathcal T$, by sampling initial state-goal pairs independently and computing the average success rate and relative cost error across tasks.

\section{Experimental evaluation}\label{sec:exp}

To evaluate \textsc{GCImOpt}, we perform experiments on four different continuous control tasks of varying complexity:
a \emph{cart-pole} system, a two-dimensional (planar) quadrotor, a three-dimensional quadrotor and a 6 degree-of-freedom (DoF) robot arm.
To model and simulate these systems, for the first three we use the \texttt{safe-control-gym} library \citep{yuan_safe-control-gym_2022}, while for the robot arm we use the \texttt{urdf2casadi} library \citep{johannessen_robot_2019} to parse a dynamic model of the robot from its URDF file.

\subsection{Dynamical systems and associated OCPs}

\paragraph{Cart-pole}
The cart-pole consists of a cart moving along a frictionless axis with a freely hinged pole that must be balanced by applying horizontal forces to the cart.
Its 4D state vector is $\bm{x} = [x, \dot{x}, \theta, \dot{\theta}]^\intercal$, describing the cart position in meters, velocity in meters per second, and the angle and angular velocity of the pole in radians and radians per second respectively.
The system's nonlinear dynamics depend on the cart and pole masses $m_c = m_p = \SI{1}{\kilo\gram}$ and pole length $l = \SI{0.5}{\meter}$.
Initial states are sampled uniformly within the bounds $[[-3, -1, 0, -1]^\intercal, [3, 1, 2\pi, 1]^\intercal]$, while goals correspond to equilibrium points with $\dot{x} = \dot{\theta} = 0$ and $\theta \in \{0, \pi\}$.
As a success indicator function for the evaluation, we require that all state variables approximate their goal values with corresponding tolerances, that is
\begin{equation}\label{eq:goal-reached}
    \textsc{GoalReached}(\bm{x}, \bm{g}) = \bigwedge\limits_{i = 1}^4 \bigg( |\bm{x}_i - \bm{g}_i| \le \varepsilon_i \bigg),
\end{equation}
where $\land$ denotes logical conjunction.
We use tolerances of $\varepsilon_x = \SI{0.05}{\meter}$, $\varepsilon_{\dot x} = \SI{0.1}{\meter\per\second}$, $\varepsilon_\theta = \SI{0.1}{\radian}$ and $\varepsilon_{\dot \theta} = \SI{0.1}{\radian\per\second}$.

\paragraph{Planar quadrotor}
The second task considers controlling a planar (2D) quadrotor, modeled as a flat plate with one vertical thruster at each edge, so there are only two control inputs $\bm{u} = [T_1, T_2]^\intercal$.
The system state is 6-dimensional, $\bm{x} = [x, \dot x, z, \dot z, \theta, \dot \theta]^\intercal$, consisting of the center of mass position and velocity, pitch angle $\theta$, and pitch rate $\dot \theta$.
The physical parameters model the Bitcraze Crazyflie 2.0.
Initial positions are sampled uniformly from a $\SI{10}{\meter}$ side box, with lower and upper state bounds $[[-5, -5, -5, -5, -\pi, -1]^\intercal, [5, 5, 5, 5, \pi, 1]^\intercal]$, while goal states always correspond to a hover at a sampled position $(x_g, z_g)$ with zero velocity and pitch.
Goal achievement is defined by tolerances on position, velocity, pitch, and pitch rate:
$\varepsilon_{\text{dist}} = \SI{5}{\centi\meter},
\varepsilon_{\text{vel}} = \SI{5}{\centi\meter\per\second},
\varepsilon_\theta = \SI{0.1}{\radian},
\varepsilon_{\dot \theta} = \SI{0.1}{\radian\per\second}$.

\paragraph{Three-dimensional quadrotor}
The third task extends the Crazyflie 2.0 quadrotor to its full three-dimensional model, with a 12-dimensional state $\bm{x} = [x, \dot x, y, \dot y, z, \dot z, \phi, \theta, \psi, p, q, r]^\intercal$, including 3D position, velocity, roll $\phi$, pitch $\theta$, yaw $\psi$ and their body frame rates.
Controls are 4-dimensional, corresponding to the quadrotor's motors.
Optimal trajectories are generated with initial and goal positions sampled from a $\SI{4}{\meter}$ box and all non-position variables initialized to zero for hover, which improves solver convergence.
Goal achievement is defined with the tolerances $\varepsilon_{\text{dist}} = \SI{5}{\centi\meter},
\varepsilon_{\text{vel}} = \SI{5}{\centi\meter\per\second},
\varepsilon_\theta = \varepsilon_\phi = \varepsilon_\psi = \SI{0.1}{\radian},
\varepsilon_{p} = \varepsilon_{q} = \varepsilon_{r} = \SI{0.1}{\radian\per\second}$.

\paragraph{Franka Emika Panda robot arm}
The final task considers controlling the Franka Emika Panda robot arm, which has 7 revolute joints.
Only 6 of the joints (A1-A6) are controlled, since the last joint (A7) does not affect the end-effector (EE) position, making the state space 6-dimensional.
The control objective is to move the EE to a specified point in space, ignoring orientation.
The EE position is obtained via forward kinematics: if $FK(\bm{x})$ yields the 3D position of the EE in configuration $\bm{x}$, the EE position (and thus the goal for the OCPs) is $\bm{y}(\bm{x}) = FK(\bm{x})$.
Controls correspond to joint angular velocities, so the dynamics simplify to $\dot{\bm{x}}(t) = \bm{u}(t)$.
Initial and goal joint angles are sampled uniformly within joint limits.
We define goal reaching success by the end-effector being within $\varepsilon = \SI{2}{\centi\meter}$ of the goal, i.e.\ $\textsc{GoalReached}(\bm{x}, \bm{g}) = |\bm{y}(\bm{x}) - \bm{g}|_2 \le \varepsilon$.

For all systems, angles are encoded as sine-cosine pairs, and when the dynamics are invariant to absolute positions, policies receive the difference between goal and current positions rather than absolute positions.
We use cost functionals that balance control effort and trajectory duration, by defining $L(\bm{x}(t), \bm{u}(t)) = \alpha |\bm{u}(t)|_2^2$ and $L_f(\bm{x}(t_f), t_f) = (1 - \alpha)t_f$ in Eq.\ \ref{eq:cost-functional}.
Thus, the cost functionals that we minimize in the generated datasets take the form
\begin{equation}\label{eq:functional}
\alpha \int_{0}^{t_f} |\bm{u}(t)|_2^2 \dd t + (1 - \alpha) t_f,
\end{equation}
where $\alpha \in [0,1]$.
Setting $\alpha=0$ minimizes the time to reach the goal, $\alpha=1$ minimizes control effort, and intermediate values interpolate between these objectives.
See Appendix~\ref{sec:app-dataset} for details on the dataset generation parameters.

\subsection{Policy training}

We train MLP policies of varying sizes for each task:
3 hidden layers for cart-pole, planar quadrotor, and robot arm, and 5 layers for the 3D quadrotor due to its higher state dimensionality.
We test hidden layer sizes of 32, 64, and 128, using the Swish activation \citep{ramachandran_swish_2017} for smooth, continuous outputs.
Policies are trained using the Adam optimizer \citep{kingma_adam_2014} with task-specific learning rates.
90\% of the trajectories are used for training and the other 10\% for validation;
we do not use a separate testing or evaluation set, since our final performance metric of interest is not regression error but closed-loop success rate and optimality gap.
We only apply goal relabeling to the training split, such that for each trajectory of $N$ grid points, we obtain $N(N+1)/2$ initial/goal state pairs.
For the 3D quadrotor, early stopping based on success rate determined the number of epochs.
We provide details on policy training in Appendix~\ref{sec:app-policy}.

\subsection{Results}\label{sec:results}

We provide the results of the experimental evaluation on closed-loop control tasks in Table\ \ref{tab:results-all-tasks}.

\begin{table}[h!]
  \caption{Final validation error $J_{\text{val}}$, success rate $s$ and average cost relative error $\delta_r$ for all tasks.
  Results are shown as mean $\pm$ standard deviation over 5 random seeds (see detailed results in Appendix~\ref{sec:app-detailed}).
  Lower $J_{\text{val}}$ and $\delta_r$ are better ($\downarrow$); higher $s$ is better ($\uparrow$).}
  \label{tab:results-all-tasks}
  \centering
  \small{
  \begin{tabular}{llrrr}
    \toprule
    Task & MLP hidden layer size & $J_{\text{val}}$ ($\downarrow$) & $s$ ($\uparrow$) & $\delta_r$ ($\downarrow$) \\
    \midrule
Cart-pole & 32 & 0.07$\pm$0.005 & 89.75$\pm$5.51\% & 21.082$\pm$3.429\% \\
 & 64 & 0.0446$\pm$0.0037 & 94.83$\pm$2.17\% & 27.252$\pm$3.444\% \\
 & 128 & 0.0321$\pm$0.0016 & 94.29$\pm$2.21\% & 29.896$\pm$12.475\% \\
\midrule
Planar quadrotor & 32 & 0.0343$\pm$0.0015 & 99.77$\pm$0.19\% & 16.888$\pm$0.932\% \\
 & 64 & 0.0146$\pm$0.0012 & 99.86$\pm$0.08\% & 7.973$\pm$0.581\% \\
 & 128 & 0.0078$\pm$0.0009 & 99.77$\pm$0.08\% & 5.282$\pm$0.838\% \\
\midrule
3D quadrotor & 32 & 0.1293$\pm$0.0688 & 28.04$\pm$18.54\% & 207.432$\pm$22.521\% \\
 & 64 & 0.0765$\pm$0.0699 & 69.83$\pm$25.55\% & 121.453$\pm$15.648\% \\
 & 128 & 0.0336$\pm$0.0094 & 97.8$\pm$2.13\% & 60.145$\pm$16.259\% \\
\midrule
Robot arm reaching & 32 & 2.6586$\pm$0.0416 & 94.76$\pm$0.64\% & 19.698$\pm$1.458\% \\
 & 64 & 2.2203$\pm$0.0398 & 98.48$\pm$0.13\% & -0.297$\pm$1.39\% \\
 & 128 & 2.3748$\pm$0.0331 & 98.71$\pm$0.21\% & -0.198$\pm$0.768\% \\
    \bottomrule
  \end{tabular}
  }
\end{table}

The results indicate that relatively small MLP policies can achieve high success rates and competitive efficiencies without extensive hyperparameter tuning.
However, there is not always a direct correlation between validation error and policy performance: in some cases, networks with higher regression errors perform better in terms of success rate or optimality gap.
This can be explained by the mismatch between the states covered by the training dataset and those visited by the policy, which appears when doing simple behavioral cloning like in our experiments.
For the cart-pole, the 128-unit MLP achieves the lowest validation error, but the 64-unit and 128-unit MLPs attain the highest average success rates, while the 32-unit MLP performs best in terms of optimality gap.
In the planar quadrotor task, all architectures achieve similar success rates, and lower validation errors correspond to smaller relative costs.

For the robot arm, success rates are consistently high, but the cost relative error decreases with larger MLPs until they become negative.
Even though $\delta_r < 0$ would imply that the trained policy is more near-optimal than the optimal policy, we observed that it is due to the non-zero tolerance of $\textsc{GoalReached}$:
when the trained policy very closely approximates the optimal control, it drives the end effector near-optimally to $\le \SI{2}{\centi\meter}$ of the goal, whereas the optimal control reaches the goal point exactly.
This goal-reaching tolerance results in the trajectory completion time (as defined in Section\ \ref{sec:eval}) being slightly earlier than that of the optimal trajectory, leading to a lower trajectory cost.
In effect, this phenomenon indicates that these robot arm policies very closely approximate the optimal controller.

The 3D quadrotor stabilization task is notably more challenging:
smaller networks failed to achieve acceptable performance, necessitating 5-layer MLPs.
Wider networks improve success rates, yet relative cost errors remain high (over 60\% on average), indicating limited policy efficiency.
Furthermore, training exhibits a counterintuitive phenomenon: as regression error decreases during the training process, success initially rises but then sharply drops.
This is likely due to the sensitivity of quadrotor dynamics, where a closer approximation of optimal controls results in aggressive maneuvers that can destabilize the system;
these destabilizations could be mitigated by covering more ``unsafe'' states in the training dataset to improve the policy's ability to correct deviations.
Overall, these findings suggest that careful task-specific tuning, more varied datasets that better cover the distribution of states visited by the policy, and incorporating domain knowledge (particularly for challenging tasks like 3D quadrotor control) would likely improve both success rates and efficiency, consistent with successful real-world deployments of similar neural network policies \citep{ferede_end--end_2024}.

In order to assess the gains in computational efficiency achieved by \textsc{GCImOpt} policies with respect to a fast trajectory optimization solver, we compare the policies' inference time and the solve time of \textsc{Fatrop}.
We show the measured times and the speedups of \textsc{GCImOpt} with respect to \textsc{Fatrop} in Table~\ref{tab:results-time}.
See Appendix~\ref{sec:app-time} for details.

\begin{table}[h!]
  \caption{Comparison of average running time between the \textsc{Fatrop} optimizer ($T_\textsc{Fatrop}$) and \textsc{GCImOpt} policies of $u$ units per hidden layer ($T_u$), all in milliseconds.
  We also report the speedup $S_u = T_\textsc{Fatrop} / T_u$.}
  \label{tab:results-time}
  \centering
  \begin{tabular}{lllllrrr}
    \toprule
    Task & $T_{\textsc{Fatrop}}$ & $T_{32}$ & $T_{64}$ & $T_{128}$ & $S_{32}$ & $S_{64}$ & $S_{128}$ \\
    \midrule
    Cart-pole & 7.530 & 0.012 & 0.011 & 0.013 & 619.857 & 669.991 & 588.896 \\
    Planar quadrotor & 32.810 & 0.013 & 0.012 & 0.014 & 2621.197 & 2724.668 & 2393.192 \\
    3D quadrotor & 84.018 & 0.013 & 0.014 & 0.015 & 6278.703 & 6188.510 & 5578.079 \\
    Robot arm reaching & 4.671 & 0.042 & 0.043 & 0.048 & 109.951 & 107.418 & 97.844 \\
    \bottomrule
  \end{tabular}
\end{table}

As shown by the comparison in inference time, \textsc{GCImOpt} policies can obtain very large speedups with respect to the fast \textsc{Fatrop} solver.
The speedups are largest in the case of the 3D quadrotor, since the high dimensionality and nonlinearity of the dynamical system makes trajectory optimization more computationally expensive.
Given that our policies are parameterized by MLPs, they have a constant computation latency;
in contrast, the execution time of an iterative trajectory optimization solver can vary depending on the initial state and goal.
The obtained speedups, together with the small size of the trained MLPs, suggest that \textsc{GCImOpt} policies could be deployed on resource-constrained controllers and achieve control frequencies rivaling or surpassing those obtained by MPC controllers.

\section{Conclusion}

In this work, we demonstrate that small neural networks trained via behavioral cloning on datasets of optimal trajectories can achieve high success rates and near-optimal goal-conditioned control across multiple tasks.
While performance generally improves with model capacity, success rates and optimality gaps do not always correlate directly with regression error.
The three-dimensional quadrotor task was the most challenging, requiring larger networks and careful tuning due to its high dimensionality and sensitivity to control errors.
In contrast, lower-dimensional tasks such as cart-pole and planar quadrotor control were solved effectively even with small MLPs.
Besides achieving good closed-loop performance, \textsc{GCImOpt} policies yield considerable reductions in control latency (up to more than $6000\times$) with respect to a fast trajectory optimization solver such as \textsc{Fatrop}.

Our approach is computationally feasible, with dataset generation and policy training achievable in under an hour on standard hardware, and readily parallelizable to scale to larger datasets.
\textsc{GCImOpt} relies on having access to an accurate system model and full observability of system states;
although these enable generating high-quality expert data for policy training, they can also result limiting in more complex applications.
Future work could address these limitations by exploring alternative policy architectures, applying the method to tasks with partial observability and sensor/actuator noise, deploying \textsc{GCImOpt} policies on real hardware, and evaluating the method on more control tasks.
In addition, although our OCP formulation can include expressive path constraints by Eq.\ \ref{eq:path-constraint}, in this work we only incorporate actuator limits and simple state bounds (such as the joint limits of the robot arm);
training constrained goal-conditioned policies is an important direction for future work.
Overall, our results show that imitating optimal trajectories is an effective method for training goal-conditioned policies that are near-optimal and computationally efficient.

\acks{
  This research was partially funded by the European Research Council under grant agreement No 101042702 Intevol-ERC2021-STG. 
  We acknowledge support by the UpnaLab research group of the Public University of Navarre (UPNA).
  This work is based on Jon Goikoetxea's Bachelor's thesis \citep{goikoetxea_learning_2025} defended in UPNA:
  we thank Patricia Yanguas, José Antonio Sanz, Daniel Aláez and Mikel Martínez-Goikoetxea for their valuable feedback during the development of the thesis.
  We also thank the anonymous L4DC 2026 reviewers for their insightful feedback and suggestions.
}
\bibliography{references}

@article{betts_survey_1998,
	title = {Survey of {Numerical} {Methods} for {Trajectory} {Optimization}},
	volume = {21},
	issn = {0731-5090, 1533-3884},
	url = {https://arc.aiaa.org/doi/10.2514/2.4231},
	doi = {10.2514/2.4231},
	language = {en},
	number = {2},
	urldate = {2025-01-24},
	journal = {Journal of Guidance, Control, and Dynamics},
	author = {Betts, John T.},
	month = mar,
	year = {1998},
	pages = {193--207},
}

@article{sanchez-sanchez_real-time_2018,
	title = {Real-{Time} {Optimal} {Control} via {Deep} {Neural} {Networks}: {Study} on {Landing} {Problems}},
	volume = {41},
	issn = {0731-5090, 1533-3884},
	shorttitle = {Real-{Time} {Optimal} {Control} via {Deep} {Neural} {Networks}},
	url = {https://arc.aiaa.org/doi/10.2514/1.G002357},
	doi = {10.2514/1.G002357},
	language = {en},
	number = {5},
	urldate = {2025-01-26},
	journal = {Journal of Guidance, Control, and Dynamics},
	author = {Sánchez-Sánchez, Carlos and Izzo, Dario},
	month = may,
	year = {2018},
	pages = {1122--1135},
}

@book{betts_practical_2010,
	title = {Practical methods for optimal control and estimation using nonlinear programming},
	publisher = {SIAM},
	author = {Betts, John T},
	year = {2010},
}

@inproceedings{mordatch_combining_2014,
	title = {Combining the benefits of function approximation and trajectory optimization.},
	volume = {4},
	booktitle = {Robotics: {Science} and {Systems}},
	author = {Mordatch, Igor and Todorov, Emo},
	year = {2014},
	pages = {23},
}

@misc{bradbury_jax_2018,
	title = {{JAX}: composable transformations of {Python}+{NumPy} programs},
	url = {http://github.com/jax-ml/jax},
	author = {Bradbury, James and Frostig, Roy and Hawkins, Peter and Johnson, Matthew James and Leary, Chris and Maclaurin, Dougal and Necula, George and Paszke, Adam and VanderPlas, Jake and Wanderman-Milne, Skye and Zhang, Qiao},
	year = {2018},
}

@article{kidger_equinox_2021,
	title = {Equinox: neural networks in {JAX} via callable {PyTrees} and filtered transformations},
	journal = {arXiv preprint arXiv:2111.00254},
	author = {Kidger, Patrick and Garcia, Cristian},
	year = {2021},
}

@inproceedings{vanroye_fatrop_2023,
	title = {{FATROP}: {A} fast constrained optimal control problem solver for robot trajectory optimization and control},
	booktitle = {2023 {IEEE}/{RSJ} {International} {Conference} on {Intelligent} {Robots} and {Systems} ({IROS})},
	publisher = {IEEE},
	author = {Vanroye, Lander and Sathya, Ajay and De Schutter, Joris and Decré, Wilm},
	year = {2023},
	pages = {10036--10043},
}

@article{andersson_casadi_2019,
	title = {{CasADi}: a software framework for nonlinear optimization and optimal control},
	volume = {11},
	journal = {Mathematical Programming Computation},
	author = {Andersson, Joel AE and Gillis, Joris and Horn, Greg and Rawlings, James B and Diehl, Moritz},
	year = {2019},
	pages = {1--36},
}

@article{ghosh_learning_2019,
	title = {Learning to reach goals via iterated supervised learning},
	journal = {arXiv preprint arXiv:1912.06088},
	author = {Ghosh, Dibya and Gupta, Abhishek and Reddy, Ashwin and Fu, Justin and Devin, Coline and Eysenbach, Benjamin and Levine, Sergey},
	year = {2019},
}

@article{liu_goal-conditioned_2022,
	title = {Goal-conditioned reinforcement learning: {Problems} and solutions},
	journal = {arXiv preprint arXiv:2201.08299},
	author = {Liu, Minghuan and Zhu, Menghui and Zhang, Weinan},
	year = {2022},
}

@book{pontryagin_mathematical_1962,
	title = {Mathematical theory of optimal processes},
	publisher = {New York: John Wiley and Sons},
	author = {Pontryagin, Lev Semenovich},
	year = {1962},
}

@article{wachter_implementation_2006,
	title = {On the implementation of an interior-point filter line-search algorithm for large-scale nonlinear programming},
	volume = {106},
	journal = {Mathematical programming},
	author = {Wächter, Andreas and Biegler, Lorenz T},
	year = {2006},
	pages = {25--57},
}

@inproceedings{kaelbling_learning_1993,
	title = {Learning to achieve goals},
	volume = {2},
	booktitle = {{IJCAI}},
	publisher = {Citeseer},
	author = {Kaelbling, Leslie Pack},
	year = {1993},
	pages = {1094--8},
}

@inproceedings{schaul_universal_2015,
	title = {Universal value function approximators},
	booktitle = {International conference on machine learning},
	publisher = {PMLR},
	author = {Schaul, Tom and Horgan, Daniel and Gregor, Karol and Silver, David},
	year = {2015},
	pages = {1312--1320},
}

@inproceedings{lynch_learning_2020,
	title = {Learning latent plans from play},
	booktitle = {Conference on robot learning},
	publisher = {PMLR},
	author = {Lynch, Corey and Khansari, Mohi and Xiao, Ted and Kumar, Vikash and Tompson, Jonathan and Levine, Sergey and Sermanet, Pierre},
	year = {2020},
	pages = {1113--1132},
}

@inproceedings{shah_gnm_2023,
	title = {Gnm: {A} general navigation model to drive any robot},
	booktitle = {2023 {IEEE} {International} {Conference} on {Robotics} and {Automation} ({ICRA})},
	publisher = {IEEE},
	author = {Shah, Dhruv and Sridhar, Ajay and Bhorkar, Arjun and Hirose, Noriaki and Levine, Sergey},
	year = {2023},
	pages = {7226--7233},
}

@inproceedings{levine_guided_2013,
	title = {Guided policy search},
	booktitle = {International conference on machine learning},
	publisher = {PMLR},
	author = {Levine, Sergey and Koltun, Vladlen},
	year = {2013},
	pages = {1--9},
}

@article{tailor_learning_2019,
	title = {Learning the optimal state-feedback via supervised imitation learning},
	volume = {3},
	number = {4},
	journal = {Astrodynamics},
	author = {Tailor, Dharmesh and Izzo, Dario},
	year = {2019},
	pages = {361--374},
}

@inproceedings{ross_reduction_2011,
	title = {A reduction of imitation learning and structured prediction to no-regret online learning},
	booktitle = {Proceedings of the fourteenth international conference on artificial intelligence and statistics},
	publisher = {JMLR Workshop and Conference Proceedings},
	author = {Ross, Stéphane and Gordon, Geoffrey and Bagnell, Drew},
	year = {2011},
	pages = {627--635},
}

@article{chi_diffusion_2023,
	title = {Diffusion policy: {Visuomotor} policy learning via action diffusion},
	journal = {The International Journal of Robotics Research},
	author = {Chi, Cheng and Xu, Zhenjia and Feng, Siyuan and Cousineau, Eric and Du, Yilun and Burchfiel, Benjamin and Tedrake, Russ and Song, Shuran},
	year = {2023},
	pages = {02783649241273668},
}

@article{reuss_goal-conditioned_2023,
	title = {Goal-conditioned imitation learning using score-based diffusion policies},
	journal = {arXiv preprint arXiv:2304.02532},
	author = {Reuss, Moritz and Li, Maximilian and Jia, Xiaogang and Lioutikov, Rudolf},
	year = {2023},
}

@article{andrychowicz_hindsight_2017,
	title = {Hindsight experience replay},
	volume = {30},
	journal = {Advances in neural information processing systems},
	author = {Andrychowicz, Marcin and Wolski, Filip and Ray, Alex and Schneider, Jonas and Fong, Rachel and Welinder, Peter and McGrew, Bob and Tobin, Josh and Pieter Abbeel, OpenAI and Zaremba, Wojciech},
	year = {2017},
}

@article{ding_goal-conditioned_2019,
	title = {Goal-conditioned imitation learning},
	volume = {32},
	journal = {Advances in neural information processing systems},
	author = {Ding, Yiming and Florensa, Carlos and Abbeel, Pieter and Phielipp, Mariano},
	year = {2019},
}

@inproceedings{codevilla_end--end_2018,
	title = {End-to-end driving via conditional imitation learning},
	booktitle = {2018 {IEEE} international conference on robotics and automation ({ICRA})},
	publisher = {IEEE},
	author = {Codevilla, Felipe and Müller, Matthias and López, Antonio and Koltun, Vladlen and Dosovitskiy, Alexey},
	year = {2018},
	pages = {4693--4700},
}

@article{ferede_end--end_2024,
	title = {End-to-end neural network based optimal quadcopter control},
	volume = {172},
	journal = {Robotics and Autonomous Systems},
	author = {Ferede, Robin and de Croon, Guido and De Wagter, Christophe and Izzo, Dario},
	year = {2024},
	pages = {104588},
}

@inproceedings{sanchez-sanchez_optimal_2016,
	title = {Optimal real-time landing using deep networks},
	volume = {12},
	booktitle = {Proceedings of the sixth international conference on astrodynamics tools and techniques, {ICATT}},
	author = {Sánchez-Sánchez, Carlos and Izzo, Dario and Hennes, Daniel},
	year = {2016},
	pages = {2493--2537},
}

@article{diehl_numerical_2011,
	title = {Numerical optimal control},
	volume = {258},
	journal = {Optimization in Engineering Center (OPTEC)},
	author = {Diehl, Moritz and Gros, Sébastien},
	year = {2011},
}

@book{rawlings_model_2017,
	title = {Model predictive control: theory, computation, and design},
	volume = {2},
	publisher = {Nob Hill Publishing Madison, WI},
	author = {Rawlings, James Blake and Mayne, David Q and Diehl, Moritz and {others}},
	year = {2017},
}

@article{rubinsztejn_neural_2020,
	title = {Neural network optimal control in astrodynamics: {Application} to the missed thrust problem},
	volume = {176},
	journal = {Acta astronautica},
	author = {Rubinsztejn, Ari and Sood, Rohan and Laipert, Frank E},
	year = {2020},
	pages = {192--203},
}

@article{yuan_safe-control-gym_2022,
	title = {Safe-{Control}-{Gym}: {A} {Unified} {Benchmark} {Suite} for {Safe} {Learning}-{Based} {Control} and {Reinforcement} {Learning} in {Robotics}},
	volume = {7},
	doi = {10.1109/LRA.2022.3196132},
	number = {4},
	journal = {IEEE Robotics and Automation Letters},
	author = {Yuan, Zhaocong and Hall, Adam W. and Zhou, Siqi and Brunke, Lukas and Greeff, Melissa and Panerati, Jacopo and Schoellig, Angela P.},
	year = {2022},
	pages = {11142--11149},
}

@article{kingma_adam_2014,
	title = {Adam: {A} method for stochastic optimization},
	journal = {arXiv preprint arXiv:1412.6980},
	author = {Kingma, Diederik P and Ba, Jimmy},
	year = {2014},
}

@inproceedings{johannessen_robot_2019,
	title = {Robot {Dynamics} with {URDF} \& {CasADi}},
	booktitle = {2019 7th {International} {Conference} on {Control}, {Mechatronics} and {Automation} ({ICCMA})},
	publisher = {IEEE},
	author = {Johannessen, Lill Maria Gjerde and Arbo, Mathias Hauan and Gravdahl, Jan Tommy},
	year = {2019},
}

@article{ramachandran_swish_2017,
	title = {Swish: a self-gated activation function},
	volume = {7},
	number = {1},
	journal = {arXiv preprint arXiv:1710.05941},
	author = {Ramachandran, Prajit and Zoph, Barret and Le, Quoc V},
	year = {2017},
	pages = {5},
}

@article{dormand_family_1980,
	title = {A family of embedded {Runge}-{Kutta} formulae},
	volume = {6},
	number = {1},
	journal = {Journal of {Computational} and {Applied} {Mathematics}},
	author = {Dormand, John R and Prince, Peter J},
	year = {1980},
	pages = {19--26},
}

@inproceedings{sprague_learning_2020,
	title = {Learning dynamic-objective policies from a class of optimal trajectories},
	booktitle = {2020 59th {IEEE} {Conference} on {Decision} and {Control} ({CDC})},
	publisher = {IEEE},
	author = {Sprague, Christopher Iliffe and Izzo, Dario and Ögren, Petter},
	year = {2020},
	pages = {597--602},
}

@article{bock_multiple_1984,
	title = {A multiple shooting algorithm for direct solution of optimal control problems},
	volume = {17},
	number = {2},
	journal = {IFAC Proceedings Volumes},
	author = {Bock, Hans Georg and Plitt, Karl-Josef},
	year = {1984},
	note = {Publisher: Elsevier},
	pages = {1603--1608},
}

@article{goikoetxea_learning_2025,
	title = {Learning efficient goal-conditioned policies by imitating optimal trajectories},
	url = {https://academica-e.unavarra.es/handle/2454/55206},
	author = {Goikoetxea, Jon},
	year = {2025},
	note = {Publisher: Public University of Navarre: Bachelor's theses repository},
}

@inproceedings{zhang_learning_2016,
	title = {Learning deep control policies for autonomous aerial vehicles with mpc-guided policy search},
	booktitle = {2016 {IEEE} {International} {Conference} on {Robotics} and {Automation} ({ICRA})},
	publisher = {IEEE},
	author = {Zhang, Tianhao and Kahn, Gregory and Levine, Sergey and Abbeel, Pieter},
	year = {2016},
	pages = {528--535},
}

@inproceedings{kahn_plato_2017,
	title = {Plato: {Policy} learning using adaptive trajectory optimization},
	booktitle = {2017 {IEEE} {International} {Conference} on {Robotics} and {Automation} ({ICRA})},
	publisher = {IEEE},
	author = {Kahn, Gregory and Zhang, Tianhao and Levine, Sergey and Abbeel, Pieter},
	year = {2017},
	pages = {3342--3349},
}

@article{da_combining_2019,
	title = {Combining trajectory optimization, supervised machine learning, and model structure for mitigating the curse of dimensionality in the control of bipedal robots},
	volume = {38},
	number = {9},
	journal = {The International Journal of Robotics Research},
	author = {Da, Xingye and Grizzle, Jessy},
	year = {2019},
	note = {Publisher: SAGE Publications Sage UK: London, England},
	pages = {1063--1097},
}

\appendix

\section{Experiment details}\label{sec:hyperparams}

All experiments, from dataset generation to policy training and evaluation, are performed on a laptop with a 13th Gen Intel i9 CPU (24 cores, 0.8-5.6 GHz) and an RTX 4090 Laptop GPU with 16 GB VRAM.

\subsection{Dataset generation}\label{sec:app-dataset}
Dataset generation is fast even with an unoptimized implementation, taking less than 12 minutes on all tasks;
generation could be accelerated further with JIT compilation, which was applied only for the 3D quadrotor.
We provide in Table~\ref{tab:datagen-summary} the dataset generation hyperparameters per task:
the number of optimized trajectories and multiple shooting grid points, the parameter $\alpha$ of the cost functional defined in Eq.~\ref{eq:functional} and the number of processes used for parallel generation.
We also report the wall-clock duration of dataset generation per task.

\begin{table}[h!]
  \caption{Parameters and duration of the dataset generation process for each task:
  number of trajectories $N_T$, number of multiple shooting grid points $N$, parameter $\alpha$ of cost functional, number of parallel processes $P$.
  Wall clock duration is shown as minutes:seconds.}
  \label{tab:datagen-summary}
  \centering
  \begin{tabular}{lrrrrr}
    \toprule
    Task & $N_T$ & $N$ & $\alpha$ & $P$ & Wall-clock duration \\
    \midrule
    Cart-pole                   & \num{20000} & 35 & 0.05 & 5 & 03:08 \\
    Planar quadrotor            & \num{20000} & 40 & 1 & 20 &   05:27 \\
    Three-dimensional quadrotor & \num{20000} & 40 & 1 & 20 &   11:23 \\
    Robot arm reaching          & \num{20000} & 35 & 0.1 & 20 & 00:19 \\
    \bottomrule
  \end{tabular}
\end{table}

\subsection{Policy training and closed-loop evaluation}\label{sec:app-policy}

Training each MLP policy, including periodic evaluation on 2000 fixed initial state-goal pairs, took about 30 minutes.
The 2000 optimal trajectories used for evaluation were generated using $N = 50$ direct multiple shooting grid points to minimize discretization error.
Table~\ref{tab:training} details the hyperparameters related to policy training, namely the minibatch size and learning rate, the number of epochs and the size of the training and validation datasets.

\begin{table}[h!]
  \caption{Summary of training hyperparameters for each task:
  minibatch size $m$, Adam learning rate $\eta$, number of training epochs $N_{\text{epochs}}$, size of training and validation splits $|\mathcal D_{\text{train}}|$ and $|\mathcal D_{\text{val}}|$.
  Note that goal relabeling is applied to the training split trajectories, yielding a large number of samples $|\mathcal D_{\text{train}}|$.}
  \label{tab:training}
  \centering
  \begin{tabular}{lrrrrr}
    \toprule
    Task & $m$ & $\eta$ & $N_{\text{epochs}}$ & $|\mathcal D_{\text{train}}|$ & $|\mathcal D_{\text{val}}|$\\
    \midrule
    Cart-pole                   & 8192 & $\num{3e-4}$ & 300 & \num{11340000} & \num{70000} \\
    Planar quadrotor            & 8192 & $\num{1e-4}$ & 200 & \num{14760000} & \num{80000} \\
    Three-dimensional quadrotor & 65536 & $\num{5e-4}$ & Early stopping & \num{14760000} & \num{80000} \\
    Robot arm reaching          & 8192 & $\num{5e-4}$ & 100 & \num{11340000} & \num{70000} \\
    \bottomrule
  \end{tabular}
\end{table}

\subsection{Inference time comparison}\label{sec:app-time}

In order to obtain the results of Table~\ref{tab:results-time}, we take $T_u$ as the average CPU inference time over $10^6$ policy forward passes.
We measure trajectory optimization solve time $T_\textsc{Fatrop}$ as the average over the 2000 initial-state/goal times sampled from the task distribution $\mathcal T$, with the same number of multiple shooting grid points as used for dataset generation (Table~\ref{tab:datagen-summary}).
We always enable JIT compilation when measuring $T_\textsc{Fatrop}$ so that the solver is compiled to an efficient C language routine with the GCC flags \texttt{-O7 -march=native}, maximizing performance as would be done in an MPC application.

\subsection{Detailed closed-loop evaluation results}\label{sec:app-detailed}

Table~\ref{tab:results-detailed} shows the results of closed-loop evaluation for 5 random seeds for each task/MLP size combination.
The random seeds are used to initialize MLP parameters and for stochastic minibatch training.
For the 3D quadrotor experiment, we select for each seed the checkpoint with the highest success rate and report its performance:
we thus effectively implement early stopping since performance degrades when training beyond this point, as discussed in Section~\ref{sec:results}.

\begin{table}[h!]
  \caption{Final validation error $J_{\text{val}}$, success rate $s$ and average cost relative error $\delta_r$ for all tasks, for 5 random seeds for each task/MLP hidden layer size ($d_\text{MLP}$) combination.}
  \label{tab:results-detailed}
  \centering
\tiny{
  \begin{minipage}{0.45\textwidth}
    
    \centering
  \begin{tabular}{llrrr}
    \toprule
    Task & $d_\text{MLP}$ & $J_{\text{val}}$ ($\downarrow$) & $s$ ($\uparrow$) & $\delta_r$ ($\downarrow$) \\
    \midrule
Cart-pole & 32 & 0.06534 & 94.4\% & 19.175\% \\
 &  & 0.06638 & 95.85\% & 22.024\% \\
 &  & 0.07293 & 84.8\% & 16.491\% \\
 &  & 0.06809 & 90.1\% & 22.132\% \\
 &  & 0.0774 & 83.6\% & 25.59\% \\
\cmidrule{2-5}
 & 64 & 0.04246 & 96.2\% & 29.945\% \\
 &  & 0.04002 & 92.6\% & 29.642\% \\
 &  & 0.04448 & 96.2\% & 25.795\% \\
 &  & 0.04667 & 96.8\% & 21.849\% \\
 &  & 0.04961 & 92.35\% & 29.029\% \\
\cmidrule{2-5}
 & 128 & 0.0322 & 90.55\% & 34.814\% \\
 &  & 0.03043 & 95.5\% & 12.976\% \\
 &  & 0.03172 & 96.1\% & 46.777\% \\
 &  & 0.03165 & 94.1\% & 30.205\% \\
 &  & 0.03475 & 95.2\% & 24.706\% \\
\midrule
Planar quadrotor & 32 & 0.03556 & 99.9\% & 16.378\% \\
 &  & 0.03197 & 99.9\% & 16.259\% \\
 &  & 0.03519 & 99.45\% & 16.011\% \\
 &  & 0.03362 & 99.85\% & 17.99\% \\
 &  & 0.03494 & 99.75\% & 17.8\% \\
\cmidrule{2-5}
 & 64 & 0.01581 & 99.9\% & 8.1\% \\
 &  & 0.01541 & 99.9\% & 8.078\% \\
 &  & 0.01505 & 99.75\% & 7.55\% \\
 &  & 0.01292 & 99.95\% & 7.319\% \\
 &  & 0.01364 & 99.8\% & 8.82\% \\
\cmidrule{2-5}
 & 128 & 0.00881 & 99.7\% & 5.206\% \\
 &  & 0.0085 & 99.7\% & 5.223\% \\
 &  & 0.00801 & 99.75\% & 6.669\% \\
 &  & 0.00697 & 99.9\% & 4.869\% \\
 &  & 0.00693 & 99.8\% & 4.444\% \\
\bottomrule

  \end{tabular}
\end{minipage}
  \begin{minipage}{0.45\textwidth}
    \centering
  \begin{tabular}{llrrr}
    \toprule
    Task & $d_\text{MLP}$ & $J_{\text{val}}$ ($\downarrow$) & $s$ ($\uparrow$) & $\delta_r$ ($\downarrow$) \\
    \midrule
3D quadrotor & 32 & 0.1361 & 54.1\% & 213.464\% \\
 &  & 0.13126 & 8.85\% & 198.342\% \\
 &  & 0.22958 & 23.65\% & 237.89\% \\
 &  & 0.03699 & 14.45\% & 211.151\% \\
 &  & 0.11245 & 39.15\% & 176.314\% \\
\cmidrule{2-5}
 & 64 & 0.1937 & 93.35\% & 116.171\% \\
 &  & 0.06484 & 79.45\% & 102.798\% \\
 &  & 0.01682 & 26.3\% & 144.431\% \\
 &  & 0.03046 & 72.1\% & 128.06\% \\
 &  & 0.07687 & 77.95\% & 115.803\% \\
\cmidrule{2-5}
 & 128 & 0.04012 & 99.35\% & 50.075\% \\
 &  & 0.01858 & 94.25\% & 86.329\% \\
 &  & 0.03586 & 99.0\% & 50.372\% \\
 &  & 0.03128 & 97.35\% & 65.773\% \\
 &  & 0.04233 & 99.05\% & 48.176\% \\
\midrule
Robot arm reaching & 32 & 2.665 & 95.15\% & 18.062\% \\
 &  & 2.63001 & 93.65\% & 19.214\% \\
 &  & 2.72684 & 95.2\% & 20.604\% \\
 &  & 2.62262 & 94.8\% & 18.885\% \\
 &  & 2.64873 & 95.0\% & 21.725\% \\
\cmidrule{2-5}
 & 64 & 2.236 & 98.4\% & -0.84\% \\
 &  & 2.16017 & 98.4\% & -1.834\% \\
 &  & 2.23723 & 98.5\% & -0.043\% \\
 &  & 2.20416 & 98.7\% & -0.675\% \\
 &  & 2.26407 & 98.4\% & 1.908\% \\
\cmidrule{2-5}
 & 128 & 2.364 & 98.7\% & -0.336\% \\
 &  & 2.35559 & 99.05\% & 0.699\% \\
 &  & 2.40519 & 98.7\% & -0.636\% \\
 &  & 2.33619 & 98.6\% & 0.446\% \\
 &  & 2.41328 & 98.5\% & -1.162\% \\
    \bottomrule
  \end{tabular}
\end{minipage}
}
\end{table}

\end{document}